\documentclass[conference]{IEEEtran}
\IEEEoverridecommandlockouts
\usepackage{cite}
\usepackage{amsmath,amssymb,amsfonts}
\usepackage{algorithmic}
\usepackage{graphicx}
\usepackage{textcomp}
\usepackage{amsmath}
\usepackage{xcolor}
\usepackage{float}
\def\BibTeX{{\rm B\kern-.05em{\sc i\kern-.025em b}\kern-.08em
    T\kern-.1667em\lower.7ex\hbox{E}\kern-.125emX}}
\begin{document}

\title{Validating transformers for redaction of text from electronic health records in real-world healthcare}

\author{\IEEEauthorblockN{1\textsuperscript{st} Zeljko Kraljevic}
\IEEEauthorblockA{\textit{Biostatistics and Health Informatics, IoPPN} \\
\textit{King's College London}\\
London, United Kingdom \\
zeljko.kraljevic@kcl.ac.uk}

\and
\IEEEauthorblockN{2\textsuperscript{nd} Anthony Shek}
\IEEEauthorblockA{\textit{Biostatistics and Health Informatics, IoPPN} \\
\textit{King's College London}\\
London, United Kingdom \\
anthony.shek@kcl.ac.uk}

\and
\IEEEauthorblockN{3\textsuperscript{rd} Joshua Au Yeung}
\IEEEauthorblockA{\textit{Department of Neurology} \\
\textit{King's College Hospital}\\
London, United Kingdom \\
j.auyeung@nhs.net}

\and
\IEEEauthorblockN{4\textsuperscript{th} Ewart Jonathan Sheldon}
\IEEEauthorblockA{\textit{Precision Medicine} \\
\textit{King's College Hospital}\\
London, United Kingdom \\
ewart.sheldon@nhs.net}

\and
\IEEEauthorblockN{5\textsuperscript{th} Haris Shuaib}
\IEEEauthorblockA{\textit{Guy’s and St Thomas’ NHS Foundation Trust}\\
London, United Kingdom \\
haris.shuaib@kcl.ac.uk}

\and
\IEEEauthorblockN{6\textsuperscript{th} Mohammad Al-Agil}
\IEEEauthorblockA{\textit{Precision Medicine} \\
\textit{King's College Hospital}\\
London, United Kingdom \\
mohammad.al-agil@nhs.net}

\and
\IEEEauthorblockN{7\textsuperscript{th} Xi Bai}
\IEEEauthorblockA{\textit{Institute of Health Informatics} \\
\textit{University College London}\\
London, United Kingdom \\
xi.bai@ucl.ac.uk}

\and
\IEEEauthorblockN{8\textsuperscript{th} Kawsar Noor}
\IEEEauthorblockA{\textit{Institute of Health Informatics} \\
\textit{University College London}\\
London, United Kingdom \\
kawsar.noor.15@ucl.ac.uk}

\and
\IEEEauthorblockN{9\textsuperscript{th} Anoop D. Shah}
\IEEEauthorblockA{\textit{Institute of Health Informatics} \\
\textit{University College London}\\
London, United Kingdom \\
a.shah@ucl.ac.uk}

\and
\IEEEauthorblockN{10\textsuperscript{th} Richard Dobson}
\IEEEauthorblockA{\textit{Biostatistics and Health Informatics, IoPPN} \\
\textit{King's College London}\\
London, United Kingdom \\
richard.dobson@kcl.ac.uk}

\and
\IEEEauthorblockN{11\textsuperscript{th} James Teo}
\IEEEauthorblockA{\textit{Department of Neurology} \\
\textit{King's College Hospital}\\
London, United Kingdom \\
jamesteo@nhs.net}
}

\maketitle

\begin{abstract}
Protecting patient privacy in healthcare records is a top priority, and redaction is a commonly used method for obscuring directly identifiable information in text. Rule-based methods have been widely used, but their precision is often low causing over-redaction of text and frequently not being adaptable enough for non-standardised or unconventional structures of personal health information. Deep learning techniques have emerged as a promising solution, but implementing them in real-world environments poses challenges due to the differences in patient record structure and language across different departments, hospitals, and countries.

In this study, we present AnonCAT, a transformer-based model and a blueprint on how deidentification models can be deployed in real-world healthcare. AnonCAT was trained through a process involving manually annotated redactions of real-world documents from three UK hospitals with different electronic health record systems and 3116 documents. The model achieved high performance in all three hospitals with a Recall of 0.99, 0.99 and 0.96. 

Our findings demonstrate the potential of deep learning techniques for improving the efficiency and accuracy of redaction in global healthcare data and highlight the importance of building workflows which not just use these models but are also able to continually fine-tune and audit the performance of these algorithms to ensure continuing effectiveness in real-world settings. This approach provides a blueprint for the real-world use of de-identifying algorithms through fine-tuning and localisation, the code together with tutorials is available on GitHub (https://github.com/CogStack/MedCAT).
\end{abstract}


\section{Introduction}
Healthcare systems contain vast amounts of unstructured health data stored in text form. While the associated structured metadata is typically anonymised according to various anonymisation guidelines (e.g. UK government guidelines and USA Health Insurance Portability and Accountability Act (HIPAA) Safe Harbour guidelines), directly identifiable information embedded in free text is harder to remove.

One commonly used method for protecting patient privacy is redaction, which involves removing or obscuring sensitive information from health records. Currently, rule-based methods such as the Protected Health Information filter (Philter)\cite{b1} and Regular Expression (RegEx) are widely used for redacting sensitive information from healthcare records. However, these methods have significant limitations, including being time-consuming to develop and maintain, and their inability to adapt to new or evolving data formats. Moreover, although these techniques achieve high recall rates, their precision is often low, leading to the concurrent redaction of clinically valuable information (e.g. eponymous conditions like Parkinson’s Disease or Smith Fracture) \cite{b2,b3,b4}.

To overcome these challenges, deep learning techniques have emerged as a promising solution for automating the redaction process. Unlike rule-based methods, deep learning algorithms can adapt to new or evolving data formats while requiring minimal manual effort. Furthermore, deep learning techniques have been shown to achieve high precision rates in redaction, minimising the risk of redacting clinically valuable information \cite{b5,b6,b7,b8}.

In a real-world deployment, implementing deep learning techniques for redacting sensitive information in healthcare records poses several challenges: the initial challenge is  generalisability primarily due to the differences in the patient record structure and language used across various departments, hospitals and countries - identifiable information in text blocks of addresses, names and official identity documents are extremely varied between geographies. After implementation, data drift which can affect both rules-based and  pre-trained deep learning models will significantly impact any redacting algorithm performance over time. This is also a bigger feature where a healthcare economy has a high socio-ethnic diversity of patients who would have a more varied structure of their personal health information (e.g. inversion of first name and last name; use of initials; multiple addresses in different countries; different types of identity documents).

The findings presented in this paper demonstrate the potential of deep learning techniques in improving the efficiency and accuracy of redaction in diverse healthcare settings. Additionally, we propose how to continually monitor and audit the performance of new and existing deep learning algorithms to ensure they remain effective and accurate and to update or fine-tune these algorithms as required. 

\section{Methods}
We present AnonCAT, which builds on the Medical Concept Annotation Toolkit (MedCAT) \cite{b9}, a widely used tool for Named Entity Recognition and Linking (NER+L) of free text from electronic health records (EHRs). This is because deidentification can be seen as NER with one additional step where we replace or remove the detected Personal/Protected Health Information (PHI) entities. It also allows us to create pipelines that can be easily deployed in hospitals and allow for easy finetuning. AnonCAT supports both the removal of PHI data and the replacement of PHI data with pseudo identifiers, we do not differentiate between these two modes of work as essentially they both depend on the ability of the model to find/detect PHI data (which is what we explore and validate).

\subsection{Ontology Creation}
\begin{figure*}[!ht]
    \center{\includegraphics [width=8cm] {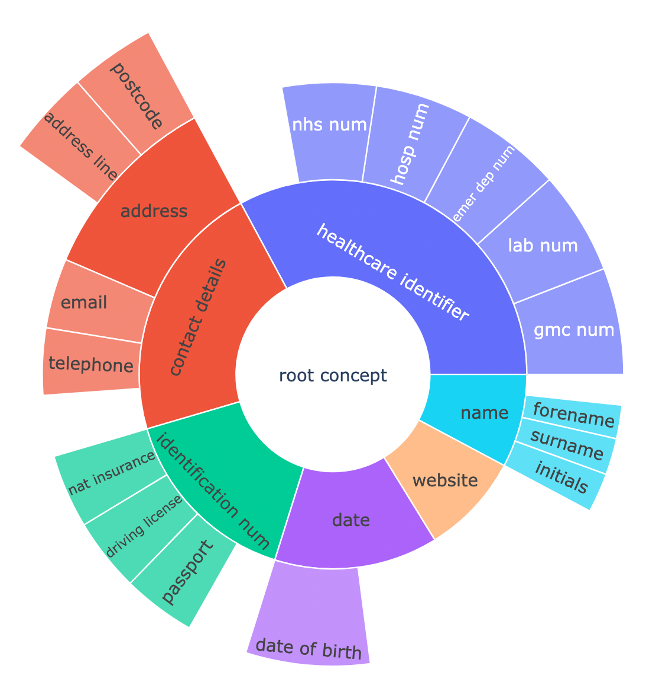}}
    \caption{Sunburst hierarchical ontology structure of terms for redaction. Five broad terms serve as the primary child nodes to which more specific terms child nodes, e.g. postcode, were connected. The terminal leaf nodes were used to annotate documents with patient data terms. Non-healthcare identifiers (green) were not prevalent enough in the text so were not used for the evaluation.}
    \label{fig:ontology)}
\end{figure*}

The UK government guidelines and USA Health Insurance Portability and Accountability Act (HIPAA) Safe Harbour guidelines were used to specify key types of data that required redaction \cite{b10}. We created an ontology from the key types (Fig. \ref{fig:ontology)}) and populated a MedCAT concept database, where each key type was represented with a MedCAT concept. The concept database was localised to the UK setting (e.g. zip code $\rightarrow$ postcode), and an ontology was chosen as it allows us to easily control the output granularity. 

\subsection{Data Annotation}
\begin{figure*}[h!]
    \centering
    \includegraphics [width=\linewidth] {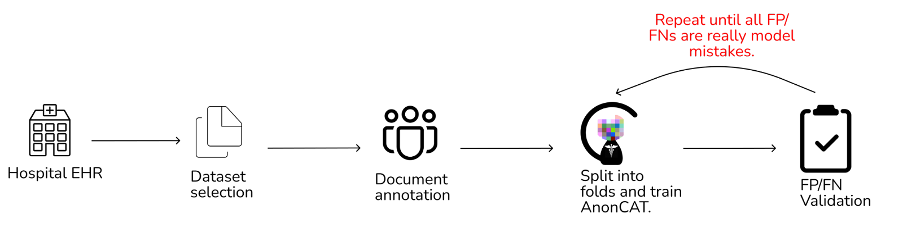}
    \caption{AnonCAT training cycle.}
    \label{fig:training_process)}
\end{figure*}

All annotations were done using MedCATtrainer \cite{b11}. Clinical annotators in each hospital (who already had visibility of the PHI data) were presented with a random document and asked to annotate all occurrences of PHI. Annotations were done by two annotators at each of the hospitals (KCH, GSTT, UCLH). To ensure standardisation between annotators and sites, we created annotation guidelines. The documents in each of the EHRs were of varying length and PHI mentions were scattered all throughout the document. As such, for very long documents, it is possible that annotators could miss a piece of PHI data. This would degrade the model performance, and also invalidate our metrics. To improve the quality of the data we used an iterative annotation approach (Fig. \ref{fig:training_process)}) to ensure all manual annotations were correct. The process is as follows:
\begin{itemize}
    \item Take a random sample of a dataset from the EHR.
    \item Clinicians / Annotators manually annotate the sampled dataset.
    \item Split the annotated dataset into 5 folds and train 5 different models, each one using a different fold for the test set.
    \item Find all false positive (FP) and false negative (FN) examples from the 5 test sets.
    \item Manually check each FP and FN, if the FP/FN is a mistake from the annotator, we fix the annotation (this creates a new dataset) and return to step 3. If the FNs and FPs represent mistakes by the model, we proceed to the next step.
    \item Once all FPs/FNs are validated and manual annotations are fixed, the process is completed.
\end{itemize}

\begin{figure*}[h]
    \centerline{\includegraphics [width=14cm]{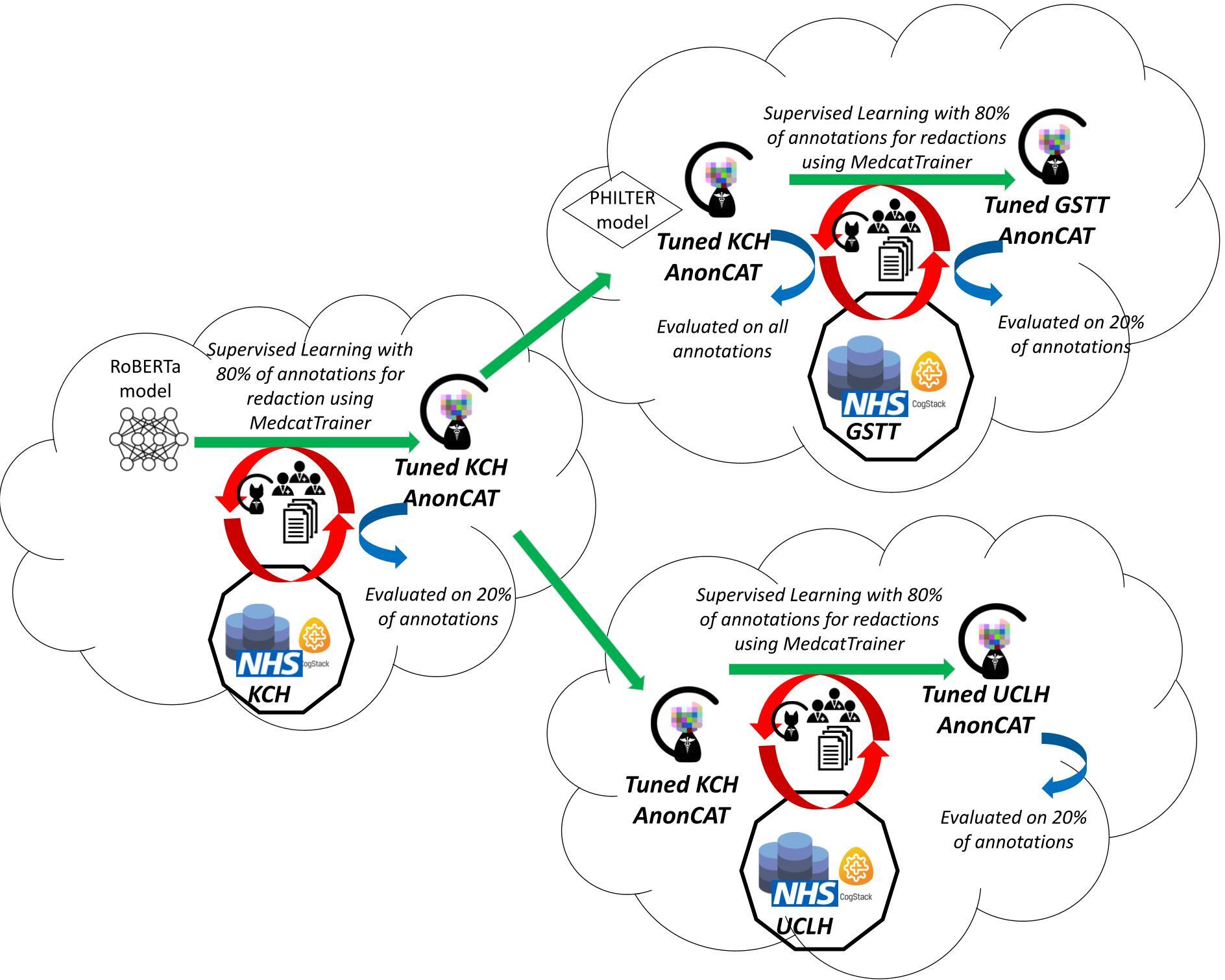}}
    \caption{The experimental setup across sites for AnonCAT.}
    \label{fig:experimental_setup)}
\end{figure*}

\subsection{Experimental Setup and Datasets}
All annotations were first pre-processed and any terms with less than 10 occurrences were removed (e.g. lab number). Next, we merged forename and surname into name, because of the observed difficulties during annotations (sometimes it was very hard to distinguish between a forename or surname).

AnonCAT uses RoBERTa-Large as the base model, in our experiments it outperformed the more clinically specialised models like BioClinicalBERT or the more general BERT model. The model was extended with a post-processing step that allows us to bias the predictions towards the positive classes. BERT-based models output the probability for each class, our modification decreases the probability of the negative class by the factor $ \lambda \in [0, 1]$, Eq. \eqref{eq1}. This often means we are increasing recall at the cost of precision.

\begin{equation}
    p_{final} (c) = 
    \begin{cases}
    p(c), & \text{if $c \neq 0$} \\
    p(c) - \lambda p(c), & \text{if $c = 0$} 
    \end{cases}
    \label{eq1}
\end{equation}
Where $p(c) \text{ is the probability of a class outputted by the model}$ and $p_{final}(c) \text{is the adjusted probability}$.

The training was done on 8 GPUs with a batch size of 4 per device, learning rate set to 4.46e-5, warm-up ratio to 0.01, weight decay to 0.14 and 10 epochs with early stopping (best results achieved around the 4th epoch). The hyperparameter tuning was done using population-based training \cite{b12} only on the train set with a 90/10 split (train/validation). On all datasets, we’ve used a train/test split of 80/20.

For the metrics, we calculated precision (P), recall (R) and F1, as well as Recall\_merged (R\_m). Recall\_merged ignores mistakes in-between concepts and only accounts for mistakes where something was supposed to be detected as PHI but was not. For example, if the token \textit{John} was detected as Address, even though it is a name, the \textit{merged} metrics will not consider this a mistake. But, if the token \textit{John} was not detected at all, that would be a mistake. This was done because, for PHI data, it is a much bigger problem if we do not detect PHI, compared to mislabeling it as another concept. When testing off-the-shelf tools for deidentification we calculated Precision\_merged (P\_m) instead of Precision, because the concepts we have do not exactly match the entity names of other tools. The merged metrics can also be understood in another way, we merge all concepts (entity names) into one called PHI, so the task becomes detecting if a token is PHI or not.

AnonCAT was tested on three different datasets: 1) A random sample of 2648 documents from the EHR at King's College Hospital NHS Foundation Trust (KCH)  2) A random sample of 328 documents from an existing recruited patient brain tumour cohort at Guy's and St Thomas' NHS Foundation Trust (GSTT); and 3) A random sample of 140 documents from the EHR at University College London Hospitals NHS Foundation Trust. To extract the datasets we used the CogStack Platform deployed at each site \cite{b13}. The statistics for the manually annotated datasets are shown in Table \ref{tbl:table1}.

\begin{table}[hp]
\caption{The number of annotations per entity and in total, as well as the total number of documents.}
\begin{center}
\begin{tabular}{|l|l|l|l|}
\hline
                            & \multicolumn{1}{c|}{\textbf{KCH}} & \multicolumn{1}{c|}{\textbf{GSTT}} & \multicolumn{1}{c|}{\textbf{UCLH}} \\ \hline
Address Line                & 2075                              & 240                                & 257                                \\ \hline
Email                       & 1740                              & 23                                 & 107                                \\ \hline
Name                        & 16407                             & 1747                               & 1743                               \\ \hline
Postcode                    & 2065                              & 208                                & 230                                \\ \hline
Emergency department Number & 2176                              & 0                                  & 0                                  \\ \hline
DOB                         & 896                               & 145                                & 112                                \\ \hline
Hosp Num                    & 1882                              & 220                                & 104                                \\ \hline
NHS Num                     & 778                               & 49                                 & 103                                \\ \hline
Initials                    & 1399                              & 168                                & 100                                \\ \hline
Telephone Num               & 1177                              & 398                                & 320                                \\ \hline
Total Annotations           & 31182                             & 3198                               & 3076                               \\ \hline
Total Documents             & 2648                              & 328                                & 140                                \\ \hline
\end{tabular}
\end{center}
\label{tbl:table1}
\end{table}

The KCH dataset (our largest one by a factor of 10) is the primary dataset where the base RoBERTa-Large model was first trained. Afterwards, the KCH model was transferred to GSTT and UCLH for further testing and fine-tuning (Fig. \ref{fig:experimental_setup)})

\subsection{Ethical approval declarations}
The project operated under London South East Research Ethics Committee (reference
18/LO/2048) approval granted to the King's Electronic Records Research Interface (KERRI) with project-specific approvals granted for NLP work on clinical coding, information governance and audit. 
For redaction validation, the GSTT corpus for patients in a directly-recruited Glioblastoma research study was used, this cohort operated under the Health Research Authority and Health and Care Research Wales (HCRW) Approval (REC reference: 18/LO/1873).
For the UCLH validation, the project was reviewed by the Trust Data Access Process for Research (DAP-R) committee, the Head of Information Governance and Data Protection Officer and approved being within the remit of service development for the overall CogStack infrastructure at UCLH. 

\section{Results}
In Table \ref{tbl:table2}, we show the performance of AnonCAT across the three hospitals, without any recall bias or modifications. The models used are as follows: 1) KCH - Roberta-Large trained on 80\% of KCH data and tested on 20\%; 2) UCLH, we fine-tuned the KCH model on 80\% of the UCLH data and tested on 20\%; and 3) GSTT, we fine-tuned the KCH model on 80\% of GSTT data and tested on 20\%. 

\begin{figure*}[!h]
    \centering
    \includegraphics[width=16cm]{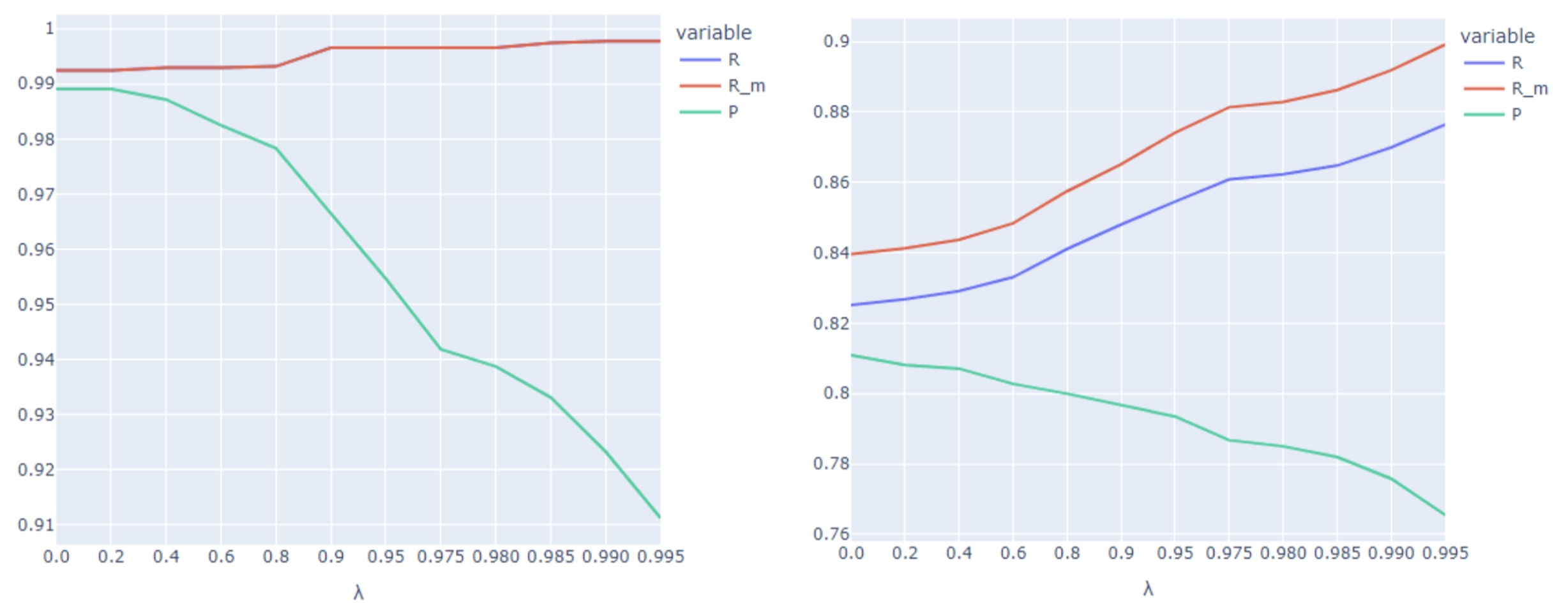}
    \caption{Results of biasing the predictions of the model towards positive classes. Left - the KCH model fine-tuned at GSTT, Right - the KCH model without any fine-tuning at GSTT ($P$ = precision, $R$ = recall, $R_m$ = recall merged). In the plot on the left, $R$ and $R_m$ overlap.}
    \label{fig:ugabuga}
\end{figure*}

\begin{table*}[h]
\caption{Performance of AnonCAT across the three datasets (P=precision, R=Recall, R\_m=Recall merged)}
\begin{center}
\begin{tabular}{|l|lll|lll|lll|}
\hline
                   & \multicolumn{3}{c|}{\textbf{KCH}}                                                                      & \multicolumn{3}{c|}{\textbf{\begin{tabular}[c]{@{}c@{}}GSTT\\ (tuned)\end{tabular}}}                   & \multicolumn{3}{c|}{\textbf{\begin{tabular}[c]{@{}c@{}}UCLH\\ (tuned)\end{tabular}}}                   \\ \hline
                   & \multicolumn{1}{c|}{\textbf{P}} & \multicolumn{1}{c|}{\textbf{R}} & \multicolumn{1}{c|}{\textbf{R\_m}} & \multicolumn{1}{c|}{\textbf{P}} & \multicolumn{1}{c|}{\textbf{R}} & \multicolumn{1}{c|}{\textbf{R\_m}} & \multicolumn{1}{c|}{\textbf{P}} & \multicolumn{1}{c|}{\textbf{R}} & \multicolumn{1}{c|}{\textbf{R\_m}} \\ \hline
Address Line       & \multicolumn{1}{l|}{0.99}       & \multicolumn{1}{l|}{0.99}       & 0.99                               & \multicolumn{1}{l|}{0.95}       & \multicolumn{1}{l|}{0.99}       & 0.99                               & \multicolumn{1}{l|}{0.80}       & \multicolumn{1}{l|}{0.96}       & 0.97                               \\ \hline
Email              & \multicolumn{1}{l|}{0.98}       & \multicolumn{1}{l|}{1.00}       & 1.00                               & \multicolumn{1}{l|}{1.00}       & \multicolumn{1}{l|}{1.00}       & 1.00                               & \multicolumn{1}{l|}{0.74}       & \multicolumn{1}{l|}{1.00}       & 1.00                               \\ \hline
Name               & \multicolumn{1}{l|}{0.99}       & \multicolumn{1}{l|}{0.99}       & 1.00                               & \multicolumn{1}{l|}{1.00}       & \multicolumn{1}{l|}{0.99}       & 0.99                               & \multicolumn{1}{l|}{0.70}       & \multicolumn{1}{l|}{0.96}       & 0.98                               \\ \hline
Postcode           & \multicolumn{1}{l|}{0.99}       & \multicolumn{1}{l|}{1.00}       & 1.00                               & \multicolumn{1}{l|}{1.00}       & \multicolumn{1}{l|}{1.00}       & 1.00                               & \multicolumn{1}{l|}{0.79}       & \multicolumn{1}{l|}{0.99}       & 0.99                               \\ \hline
Emergency dept Num & \multicolumn{1}{l|}{0.99}       & \multicolumn{1}{l|}{0.97}       & 1.00                               & \multicolumn{1}{l|}{NA}         & \multicolumn{1}{l|}{NA}         & NA                                 & \multicolumn{1}{l|}{NA}         & \multicolumn{1}{l|}{NA}         & NA                                 \\ \hline
DOB                & \multicolumn{1}{l|}{0.99}       & \multicolumn{1}{l|}{0.98}       & 1.00                               & \multicolumn{1}{l|}{1.00}       & \multicolumn{1}{l|}{1.00}       & 1.00                               & \multicolumn{1}{l|}{0.90}       & \multicolumn{1}{l|}{1.00}       & 1.00                               \\ \hline
Hosp Num           & \multicolumn{1}{l|}{1.00}       & \multicolumn{1}{l|}{1.00}       & 1.00                               & \multicolumn{1}{l|}{1.00}       & \multicolumn{1}{l|}{1.00}       & 1.00                               & \multicolumn{1}{l|}{0.78}       & \multicolumn{1}{l|}{1.00}       & 1.00                               \\ \hline
NHS Num            & \multicolumn{1}{l|}{1.00}       & \multicolumn{1}{l|}{0.98}       & 1.00                               & \multicolumn{1}{l|}{1.00}       & \multicolumn{1}{l|}{1.00}       & 1.00                               & \multicolumn{1}{l|}{0.94}       & \multicolumn{1}{l|}{0.94}       & 0.99                               \\ \hline
Initials           & \multicolumn{1}{l|}{0.98}       & \multicolumn{1}{l|}{0.94}       & 0.95                               & \multicolumn{1}{l|}{1.00}       & \multicolumn{1}{l|}{0.97}       & 0.97                               & \multicolumn{1}{l|}{0.81}       & \multicolumn{1}{l|}{0.81}       & 0.81                               \\ \hline
Telephone Num      & \multicolumn{1}{l|}{0.98}       & \multicolumn{1}{l|}{1.00}       & 1.00                               & \multicolumn{1}{l|}{0.95}       & \multicolumn{1}{l|}{0.99}       & 0.99                               & \multicolumn{1}{l|}{0.83}       & \multicolumn{1}{l|}{0.96}       & 0.96                               \\ \hline
\end{tabular}
\end{center}
\label{tbl:table2}
\end{table*}
\begin{table}[H]
\caption{Performance of Philter and the AnonCAT KCH model without any finetuning at GSTT. We are only calculating merged metrics as the entities in Philter do not match exactly the entities in our concept database.}
\begin{center}
\begin{tabular}{|l|ll|ll|}
\hline
                   & \multicolumn{2}{c|}{\textbf{\begin{tabular}[c]{@{}c@{}}AnonCAT\\ (untuned)\end{tabular}}} & \multicolumn{2}{c|}{\textbf{Philter}}                                   \\ \hline
                   & \multicolumn{1}{c|}{\textbf{P\_m}}          & \multicolumn{1}{c|}{\textbf{R\_m}}          & \multicolumn{1}{c|}{\textbf{P\_m}} & \multicolumn{1}{c|}{\textbf{R\_M}} \\ \hline
Address Line       & \multicolumn{1}{l|}{0.79}                   & 0.99                                        & \multicolumn{1}{l|}{0.37}          & 0.79                               \\ \hline
Email              & \multicolumn{1}{l|}{0.79}                   & 0.94                                        & \multicolumn{1}{l|}{0.37}          & 1                                  \\ \hline
Name               & \multicolumn{1}{l|}{0.79}                   & 0.99                                        & \multicolumn{1}{l|}{0.37}          & 0.99                               \\ \hline
Postcode           & \multicolumn{1}{l|}{0.79}                   & 1                                           & \multicolumn{1}{l|}{0.37}          & 0.51                               \\ \hline
Emergency dept Num & \multicolumn{1}{l|}{0.79}                   & NA                                          & \multicolumn{1}{l|}{NA}            & NA                                 \\ \hline
DOB                & \multicolumn{1}{l|}{0.79}                   & 0.74                                        & \multicolumn{1}{l|}{0.37}          & 1                                  \\ \hline
Hosp Num           & \multicolumn{1}{l|}{0.79}                   & 0.52                                        & \multicolumn{1}{l|}{0.37}          & 1                                  \\ \hline
NHS Num            & \multicolumn{1}{l|}{0.79}                   & 1                                           & \multicolumn{1}{l|}{0.37}          & 1                                  \\ \hline
Initials           & \multicolumn{1}{l|}{0.79}                   & 0.67                                        & \multicolumn{1}{l|}{0.37}          & 0.83                               \\ \hline
Telephone Num      & \multicolumn{1}{l|}{0.79}                   & 0.70                                        & \multicolumn{1}{l|}{0.37}          & 0.50                               \\ \hline
\end{tabular}
\end{center}
\label{tbl:table3}
\end{table}

Next, in Table \ref{tbl:table3} we show the performance on the GSTT dataset for two different models. AnonCAT KCH, we test the KCH model on the full GSTT dataset without any fine-tuning, and Philter an off-the-shelf tool showing state-of-the-art performance on de-identification of clinical notes \cite{b1}. The entities in Philter do not match exactly the entities we have in our concept database, so we only calculate the merged metrics for both Precision and Recall.

Lastly, in Fig. \ref{fig:ugabuga} we also test does biasing the model towards predicting the positive classes work, and what is the tradeoff between precision and recall. We do the test at GSTT and check the performance for two different models - AnonCAT trained only on the KCH.

\section{Discussion}
The findings of this study suggest that with appropriate fine-tuning and maintenance, deep learning techniques offer an adaptable solution for improving the accuracy and efficiency of the redaction of EHR data. This lays out a route for a foundational redaction NLP model that can be rapidly localised to a healthcare environment thereby reducing the rate of under-redaction (due to the uniqueness of patient data in new hospitals).

We show that as few as 150-300 documents are needed to fine-tune a redaction model on a new dataset, with a higher performance the more documents available for fine-tuning. We also note that during our iterative approach where we check for disagreements between the model and annotators (but also more generally validate the annotations), most (\>90\%) of mistakes are annotators overlooking a piece of PHI data, and rarely wrongly annotating something. As such, we argue that our iterative approach is a more efficient way of creating a well-annotated dataset, than the more standard double annotation approach, in cases where finding the entities is difficult, but annotating them correctly is easy. We plan to explore this in more detail in future work.

Table \ref{tbl:table3} shows that state-of-the-art off-the-shelf tools, as well as transformer-based models, require some fine-tuning when applied to a new dataset if we want to achieve a Recall of 95\%+ (as shown in Table \ref{tbl:table2}) We argue that a deep learning approach is easier to maintain and upgrade, as the only requirement for adapting it to a new dataset is annotating 150-300 documents and running the fine-tuning. In fact, the very task of annotating for redaction is already being done by information governance auditors manually for records release in subject access requests, so a minor adaptation of the workflow is all that is needed to build resilience in redaction performance.

\section{Conclusion}
We demonstrate that a Transformer-based Deep Learning approach to effective and efficient text redaction can be achieved through localised fine-tuning of a foundation model. This use of localised efficient fine-tuning would also mitigate against performance deterioration of a foundation model when deployed in closed private datasets from different underlying EHRs. 
Our workflow can be easily incorporated into existing activity when the hospital’s auditors redact text, in this way normal hospital information governance activity would generate validation metrics while continually fine-tuning models. The application of our method can significantly enhance patient privacy protection in healthcare organisations and contribute to the advancement of an efficient safe life cycle of healthcare data and NLP model management \cite{b14,b15}.

\section{Acknowledgements}
This work was made possible by funding received from NHS AI Lab, National Institutes of Health Research, Health Data Research UK, Guy’s \& St Thomas’ NHS Foundation Trust, NIHR Applied Research Collaborative South London, Guy’s and St Thomas’ Foundation, the Radiological Research Trust, and a National Institute for Health Research Doctoral Research Fellowship (ref. 20647). Special thanks as well to the patients of the KERRI committee.

R.D. is supported by the following: (1) NIHR Biomedical Research Centre at South London and Maudsley NHS Foundation Trust and King’s College London, United Kingdom; (2) Health Data Research UK; (3) the NIHR University College London Hospitals Biomedical Research Centre; (4) the UK Research and Innovation London Medical Imaging and Artificial Intelligence Centre for Value Based Healthcare; and (5) the NIHR Applied Research Collaboration South London (NIHR ARC South London) at King’s College Hospital NHS Foundation Trust.
ADS is supported by a postdoctoral fellowship from THIS Institute, NIHR (AI\_AWARD01864 and COV-LT-0009), UKRI (Horizon Europe Guarantee for DataTools4Heart) and British Heart Foundation Accelerator Award (AA/18/6/24223).

\end{document}